\documentclass[letterpaper]{article} 
\usepackage{aaai24}  
\usepackage{times}  
\usepackage{helvet}  
\usepackage{courier}  
\usepackage[hyphens]{url}  
\usepackage{graphicx} 
\usepackage{natbib}  
\usepackage{caption} 
\usepackage{algorithm}
\usepackage{algorithmic}

\usepackage{newfloat}
\usepackage{listings}
\DeclareCaptionStyle{ruled}{labelfont=normalfont,labelsep=colon,strut=off} 
\floatstyle{ruled}
\newfloat{listing}{tb}{lst}{}
\floatname{listing}{Listing}
\usepackage{booktabs}
\usepackage{color}
\usepackage{multirow}
\usepackage{amsfonts}
\usepackage{pifont}
\usepackage{amsmath}

\title{Hierarchical IoU Tracking based on Interval}
\author{
    Yunhao Du$^{1}$, Zhicheng Zhao$^{1,2,3}$, Fei Su$^{1,2,3}$
}
\affiliations{
    $^1$The school of Artificial Intelligence, Beijing University of Posts and Telecommunications \\
    $^2$Beijing Key Laboratory of Network System and Network Culture, China\\
    $^3$Key Laboratory of Interactive Technology and Experience System Ministry \\of Culture and Tourism, Beijing, China\\
}

\usepackage{bibentry}

\begin{document}


\twocolumn[{
    \renewcommand\twocolumn[1][]{#1}
    \maketitle
    \vspace{-1cm}
    \begin{center}
        \centering
        \includegraphics[width=.95\textwidth]{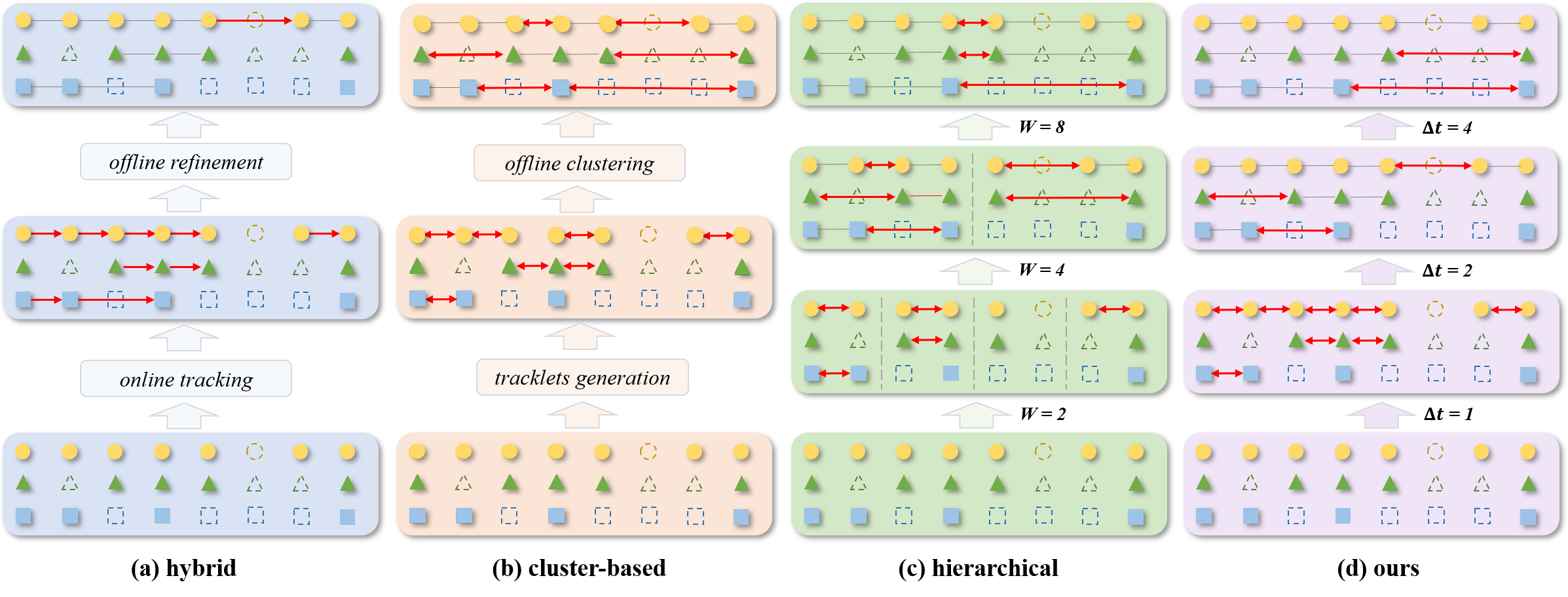}
        \captionof{figure}{
            \textbf{The comparison among different offline tracking frameworks.}
            We construct a sequence with eight frames and three targets as example, where the dashed ones represent missing detections.
            (a) Current dominant hybrid methods first track targets in an online manner, 
            and then refine trajectories with interpolation and global association.
            (b) Cluster-based methods first generate reliable tracklets and then model tracklet association as the graph partition problem for clustering.
            (c) Previous hierarchical solutions iteratively match neighboring tracklets with an increasing window size $W$.
            (d) Our framework also follows the hierarchical paradigm and gradually increases the maximum tracklet interval $\Delta t$ to ensure the purity of results.
        }
        \label{fig_compare}
    \end{center}
}]

\begin{abstract}
    Multi-Object Tracking (MOT) aims to detect and associate all targets of given classes across frames.
    Current dominant solutions, e.g. ByteTrack and StrongSORT++, follow the hybrid pipeline, 
    which first accomplish most of the associations in an online manner, and then refine the results using offline tricks such as interpolation and global link.
    While this paradigm offers flexibility in application, the disjoint design between the two stages results in suboptimal performance.
    In this paper, we propose the \textbf{H}ierarchical \textbf{I}oU \textbf{T}racking framework, dubbed \textbf{HIT},
    which achieves unified hierarchical tracking by utilizing tracklet intervals as priors.
    To ensure the conciseness, only IoU is utilized for association,
    while discarding the heavy appearance models, tricky auxiliary cues, and learning-based association modules.
    We further identify three inconsistency issues regarding target size, camera movement and hierarchical cues, 
    and design corresponding solutions to guarantee the reliability of associations.
    Though its simplicity, our method achieves promising performance on four datasets, i.e., MOT17, KITTI, DanceTrack and VisDrone,
    providing a strong baseline for future tracking method design.
    Moreover, we experiment on seven trackers and prove that HIT can be seamlessly integrated with other solutions, 
    whether they are motion-based, appearance-based or learning-based.
    Our codes will be released at \url{https://github.com/dyhBUPT/HIT}.

\end{abstract}

\section{Introduction}

Multi-Object Tracking (MOT), which involves visually distinguishing the identity of multiple objects in a scene and creating their trajectories,
is a fundamental yet crucial vision task, imperative to address numerous problems in areas such as surveillance, robotics, autonomous driving and biology.
It is commonly confronted with challenges including occlusions, missing detections, localization errors, 
non-linear motion patterns and long-term associations, necessitating further optimization efforts.

Current MOT methods follow in principle the predominant \textit{hybrid} paradigm \cite{zhang2022bytetrack, du2023strongsort} (\textcolor{red}{Fig.\ref{fig_compare}(a)}),
which elaborate online algorithms to obtain trajectories and introduce offline post-processing procedures for refinement 
such as linear interpolation and global association \cite{du2021giaotracker}.
Despite their impressive performance and flexible applicability, the inherent unreliability of online algorithms undermines the overall effectiveness.
To solve this problem, some other works formulate MOT as a two-stage clustering problem \cite{wang2019exploit, dai2021learning}, 
named \textit{cluster-based} paradigm (\textcolor{red}{Fig.\ref{fig_compare}(b)}).
They first generate short tracklets with low ambiguity using strict spatio-temporal and appearance constraints,
and then cluster them based on graph partition \cite{kumar2015multiple} or iterative proposals.
However, cluster-based methods require distinct designs tailored to different stages and timespans, which limits the usability and scalability.

Pure \textit{hierarchical} pipeline (\textcolor{red}{Fig.\ref{fig_compare}(c)}) addresses it by employing a unified design across different hierarchies,
enabling the use of a single model or algorithm for both short-term and long-term associations simultaneously \cite{cetintas2023unifying}.
In this paradigm, a set of exponentially expanding non-overlapping temporal windows are established,
where each hierarchy performs association exclusively within its corresponding windows.
Nevertheless, the design of predefined windows fail to consider the variations in inherent reliability among different trajectories.

In this paper, we propose a new hierarchical tracking framework that uses ``tracklet interval'' as the hierarchical basis instead of ``temporal window''.
As shown in \textcolor{red}{Fig.\ref{fig_compare}(d)}, in the first hierarchy ($\Delta t=1$), only detections from adjacent frames are associated.
Then in the second hierarchy ($\Delta t=2$), matching with a two-frame interval is allowed, thus tolerating one missing detection.
Similarly, higher hierarchies facilitate longer-term tracking.
Essentially, this design grants higher priority to high-quality tracklets, thereby ensuring the purity of results.
We simply employ Kalman Filter for motion prediction and IoU for association in all hierarchies,
and prove the superiority of the ``tracklet interval'' strategy over ``temporal window'' on MOT17 \cite{milan2016mot16} and KITTI \cite{geiger2013vision}.
However, three inconsistency issues are observed in our framework as follows:
\begin{itemize}
    \item \textbf{Inconsistent target size:} Given fixed pixel-level detection errors, small boxes usually exhibit lower IoU with ground truth compared to large boxes.
    \item \textbf{Inconsistent camera movement:} Camera movement scales tend to vary across different sequences.
    \item \textbf{Inconsistent hierarchical cues:} In the first hierarchy, each tracklet only contains one box, resulting in no motion information.
    Nevertheless, multiple boxes are available for motion prediction in higher hierarchies.
\end{itemize}
In conclusion, the aforementioned issues make it challenging to use unified algorithm and hyperparameters for all targets, sequences and hierarchies.
To overcome these problems, specific optimization strategies are designed to achieve unified \textit{hierarchical IoU tracking}, named \textit{HIT}.

Though its simplicity, our HIT achieves promising performance on various datasets, i.e., MOT17, KITTI, DanceTrack and VisDrone,
positioning it as a strong baseline for future tracker designs.
Specifically, we obtain the same HOTA and lower IDSW compared with StrongSORT++ \cite{du2023strongsort} on KITTI 
without using appearance features and the CMC (camera motion compensation) module.

Furthermore, by conducting experiments on seven other motion-based, appearance-based and learning-based trackers,
we demonstrate that HIT can be integrated with existing trackers for trajectory recombination and refinement, and improve the performance significantly.
This highlights the potential of HIT to serve as a new post-processing algorithm in application.

\begin{figure*}
    \centering
    \includegraphics[width=.9\textwidth]{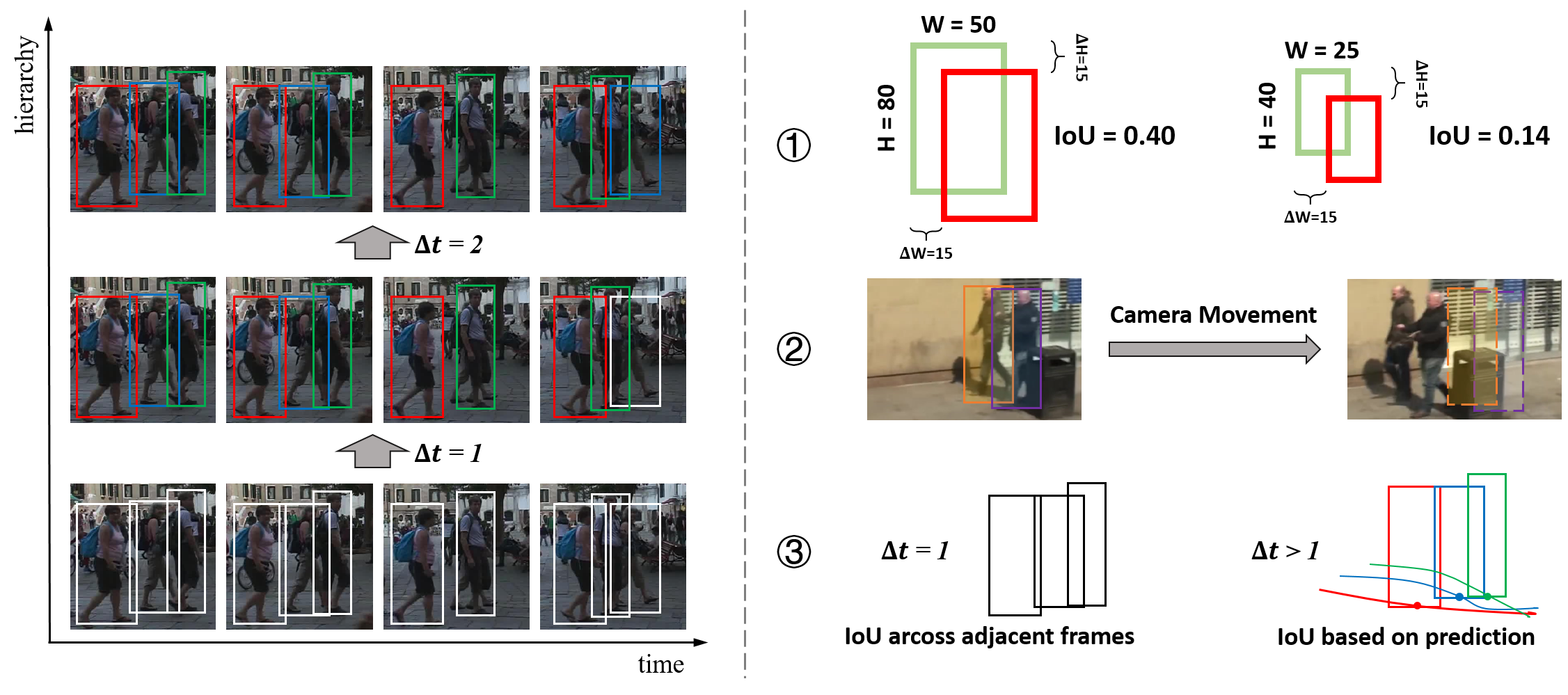}
    \captionof{figure}{
        \textbf{The illustration of our framework and three inconsistency issues.}
        \textbf{Left:} We illustrate our pipeline with a simple example with 4 frames and 3 targets.
        After the first hierarchy ($\Delta t=1$), all adjacent detections are associated.
        Then the second hierarchy ($\Delta t=2$) further identifies the missed association.
        \textbf{Right:}  \ding{172} illustrates the ``inconsistent target size'' issue, 
        in which smaller boxes tend to have lower IoU for given localization errors.
        \ding{173} shows camera movement will cause boxes misalignment across frames, which is named ``inconsistent camera movement''.
        \ding{174} clarifies the ``inconsistent hierarchical cues'', where the first hierarchy can only utilize overlap information of adjacent boxes, 
        while higher hierarchies can incorporate motion information.
    }
    \label{fig_framework}
\end{figure*}

\section{Related Work}

\subsection{Online Tracking}

In recent years, heuristic online trackers have dominated the MOT community.
SORT \cite{bewley2016simple} employed Kalman Filter \cite{1960A} for motion prediction, which later became the foundation for other works.
DeepSORT \cite{wojke2017simple} improved it by introducing extra appearance features and proposed the cascade matching algorithm.
Recently, StrongSORT \cite{du2023strongsort} upgraded DeepSORT with various advanced tricks.
ByteTrack \cite{zhang2022bytetrack} trained a powerful detector YOLOX \cite{ge2021yolox} and pioneered the use of low-confidence detection boxes.
OC-SORT \cite{cao2023observation} rethought the role of Kalman Filter in SORT and proposed three observation-centric techniques for stable association.
BoT-SORT \cite{aharon2022bot} updated the setting of Kalman Filter states and elaborated the design of CMC (camera movement compensation) module.
Hybrid-SORT \cite{yang2024hybrid} improved from OC-SORT and ByteTrack, and incorporated two weak cues, i.e. confidence and height of boxes, to compensate for strong cues.

Currently, learning-based trackers have experienced rapid development.
TransTrack \cite{sun2020transtrack} and TrackFormer \cite{meinhardt2022trackformer} proposed to employ DETR \cite{carion2020end} for joint detection and association learning,
which employed ``track query'' to ensure consistent target information to maintain tracklets.
MOTR \cite{zeng2022motr} presented a fully end-to-end MOT framework, requiring no heuristic procedures such as NMS and extra matching.
Subsequent works further focused on improving it in terms of detection \cite{zhang2023motrv2, yu2023motrv3, yan2023bridging}, 
long-term modeling \cite{gao2023memotr}, occlusion robustness \cite{fu2023denoising} and description understanding \cite{yu2023generalizing, wu2023referring}.

Though those methods were designed for online tracking, many of them utilized extra offline post-processes to refine trajectories in inference.
We claim that this hybrid pipeline results in suboptimal performance and thus commit to design a unified offline framework.

\subsection{Offline Tracking}

Most pure offline trackers followed the two-stage cluster-based paradigm, which first generated reliable tracklets based on spatio-temporal and appearance cues,
and then constructed a tracklet graph for clustering.
TAT \cite{shen2018tracklet} proposed a network flow association approach by formulating MOT as the bi-level optimization problem.
TNT \cite{wang2019exploit} constructed the TrackletNet to model location and appearance information jointly and associated tracklets using the graph partition approach.
TPM \cite{peng2020tpm} developed tracklet matching planes to resolve association confusions caused by noisy or missing detections.
DTA \cite{zhang2020long} drew inspiration from MHT \cite{kim2015multiple} to build hypothesis trees to represent multiple potential trajectories simultaneously.
LPC \cite{dai2021learning} designed an iterative graph clustering strategy for proposal generation and employed GCN \cite{kipf2016semi} to score these proposals.
FCG \cite{girbau2022multiple} fused tracklets in consecutive lifted frames in a cascade manner.

While these methods have demonstrated excellent performance, they require distinct algorithms for the two steps, i.e., tracklet generation and tracklet clustering.
Moreover, for tracklet clustering, some multi-stage solutions necessitate the use of different modules for each stage.
Recently, SUSHI \cite{Cetintas_2023_CVPR} presented a learning-based hierarchical framework, which utilized unified designs for all hierarchies.
However, it relied on temporal windows to partition different hierarchies, without considering the inherent information of tracklets.
In this work, we propose to use the intrinsic reliability cues of tracklets, i.e. tracklet intervals, to realize the hierarchical framework for better stability. 

\section{Method}

In this section, we will first present the overall hierarchical framework of our method HIT.
Then, three consistency designs are proposed to tackle the corresponding inconsistent problems.
Finally, we will introduce how to integrate HIT with other trackers.

\subsection{Framework}

Fig.\ref{fig_framework} (left) illustrates our hierarchical framework.
Given $N_1$ input detections $\cal D$ across all $T$ frames of a sequence, the initial tracklet set is constructed by treating each detection as one tracklet.
Then, in the $l$-th hierarchy ($l \ge 1$), $N_{l}$ tracklets $\mathcal{T}^l = \{\mathcal{T}_i^l\}_{i=1}^{N_l}$ 
are associated to $N_{l+1}$ longer tracklets $\mathcal{T}^{l+1} = \{\mathcal{T}_i^{l+1}\}_{i=1}^{N_{l+1}}$ ($N_{l} \ge N_{l+1}$).
Each tracklet is formulated as follows:
\begin{equation}
    \mathcal{T}_i^{l} = \{ \tau_{t}^{l,i} \}_{t=T_{min}^{l,i}}^{T_{max}^{l,i}},
\end{equation}
where $\tau_t^{l,i}$ is the detection box at frame $t$ and $T_{min}^{l,i}$ and $T_{max}^{l,i}$ are the minimum and maximum of frame indices of $\mathcal{T}_i^{l}$.

A set of tracklet interval thresholds $\mathcal{I} = \{\Delta t^l\}_{l=1}^L$ are preset. 
In the $l$-th hierarchy, only tracklet pairs $\{\mathcal{T}_i^{l}, \mathcal{T}_j^{l}\}$ with intervals smaller than the threshold are considered for association,
i.e., $0 < T_{min}^{l,j} - T_{max}^{l,i} \le \Delta t^l$.
For association, bidirectional motion prediction is performed with Kalman Filter,
and the matching similarity is computed as the IoU between the true boxes and predicted locations of tracklet pairs as in previous works \cite{zhang2022bytetrack, cao2023observation}.
A unified matching threshold $\Delta o$ is utilized for all hierarchies, and tracklets are associated by employing Hungarian algorithm \cite{kuhn1955hungarian}.
By iteratively performing the aforementioned association process from hierarchy 1 to $L$ with an increasing $\Delta t^l$, we ultimately obtain $N_{L+1}$ output trajectories.

\subsection{Consistency Designs}

Despite the effectiveness of our framework, three types of inconsistency issues are identified, as shown in Fig.\ref{fig_framework} (right).

\subsubsection{Inconsistent target size}

IoU is widely used for tracklet association and metric evaluation in MOT.
However, we find that small boxes tend to have lower IoU than large boxes.
For example, given localization errors of 15 pixels in both horizontal and vertical directions,
the IoU with GT is 0.4 for the box of size 80$\times$50, while is 0.14 for the box of size 40$\times$25.
This makes it difficult to use the same matching threshold for all targets.
To solve this problem, we propose the \textbf{consistent-IoU} to expand small boxes before calculating IoU.
Specifically, given two boxes $b_i = (x_i,y_i,w_i,h_i)$ and $b_j = (x_j,y_j,w_j,h_j)$,
we expand them with a ratio $r_{i,j}$ if $w_i < W$ and $w_j < W$ as follows:
\begin{equation} 
    \begin{aligned}
        \hat w_i &= r_{i,j} \cdot w_i, \ \ \  \hat h_i = r_{i,j} \cdot h_i, \\
        \hat w_j &= r_{i,j} \cdot w_j, \ \ \  \hat h_j = r_{i,j} \cdot h_j, \\
        r_{i,j} &= \sqrt{e^{\tau * {W \over w_i}} \cdot e^{\tau * {W \over w_j}}},
    \end{aligned}
\end{equation}
where $W$ is the preset threshold and $\tau = 0.2$ is the scaling factor.
Then the expanded boxes $\hat b_i = (x_i, y_i, \hat w_i \hat h_i)$ and $\hat b_j = (x_j, y_j, \hat w_j, \hat h_j)$ are utilized to compute IoU for association.
Note that for large boxes, we simply use the raw IoU.

\subsubsection{Inconsistent camera movement}

Our IoU-based framework highly relies on the motion information of targets across frames.
However, the scale of camera movement varies in different sequences, resulting in differences in inter-frame target overlaps.
To compensate for this gap, we propose the \textbf{consistent-camera} method to estimate the camera movement without using visual cues.
Firstly, for the $k$-th sequence, the first hierarchical association with $\Delta t = 1$ is performed.
Then we calculate the average IoU $O_k$ of all matched detection pairs as the measure of camera movement of this sequence.
If $O_k$ is smaller than the threshold $\Delta O$, the sequence is identified as having significant camera movement.
In this case, according to FOR \cite{nasseri2023online}, the degree of camera movement $(\Delta X_t, \Delta Y_t)$ at frame $t$ can be estimated 
by calculating the average distance of all associated detection pairs $\{b^t_i, b^{t+1}_i\}_{i=1}^{N_t}$ as follows:
\begin{equation}
    \begin{aligned}
        \Delta X_t &= {1 \over N_t} \sum_{i=1}^{N_t} (x_i^{t+1} - x_i^t), \\
        \Delta Y_t &= {1 \over N_t} \sum_{i=1}^{N_t} (y_i^{t+1} - y_i^t). \\
    \end{aligned}
\end{equation}
Finally, $(\Delta X_t, \Delta Y_t)$ is used to compensate for camera movement in all hierarchies as in previous CMC-based solutions \cite{du2023strongsort}.
Note that for sequences with negligible camera movement, we don't apply the compensation.

\subsubsection{Inconsistent hierarchical cues}

As previous mentioned, motion prediction is performed before calculating IoU for association.
Nevertheless, for the first hierarchy with $\Delta t=1$, each tracklet has a length of only one, resulting in insufficient temporal information for motion estimation.
This usually leads to unreliable associations, especially when two trajectories cross over.
To tackle this inconsistent puzzle, we propose the \textbf{consistent-motion} strategy to equip the first hierarchy with motion cues.
Similarly to before, the first hierarchical association is firstly performed purely based on inter-frame IoU.
For each box, the matched boxes of preceding frames can be used for motion estimation in subsequent frames, and vice versa.
Thus, all tracklets across all hierarchies contain temporal information, benefiting the unified hierarchical pipeline.

\begin{figure}[htbp]
    \centering
    \includegraphics[width=.4\textwidth]{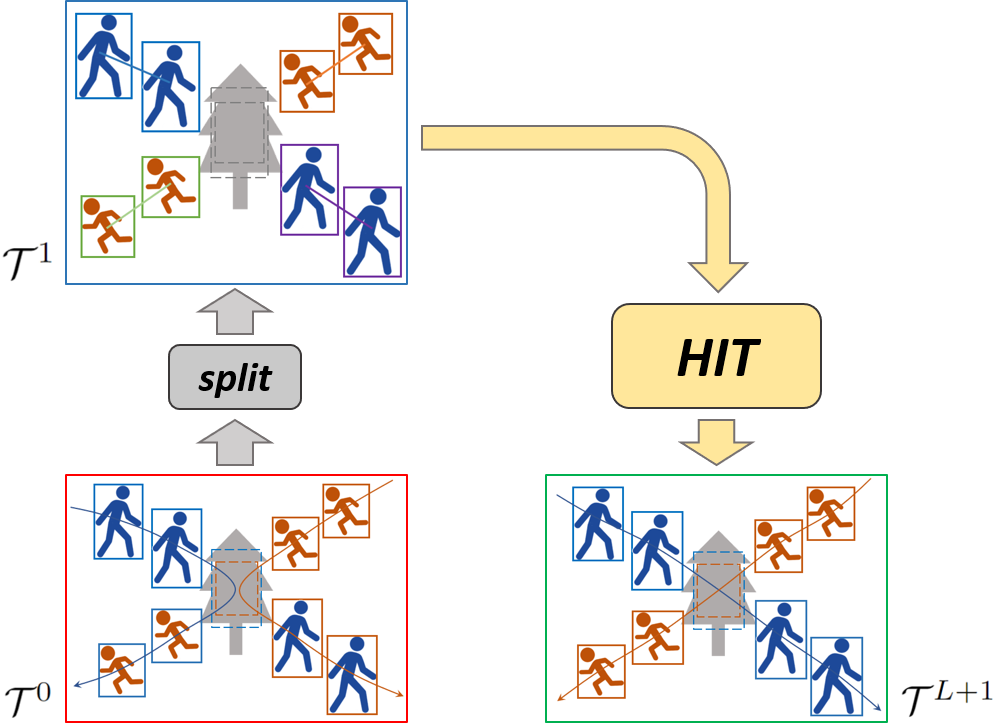}
    \captionof{figure}{
        \textbf{The illustration of integrating HIT with another tracker.}
        In this example, two trajectories are occluded and switch IDs.
        In our pipeline, they are first split into four short tracklets based on continuity and then recombined by HIT.
    }
    \label{fig_integration}
\end{figure}

\subsection{Integration with other Trackers}

As described previously, HIT utilizes detections to initialize tracklets and performs hierarchical associations.
Thus, an intriguing question arises: \textit{can we use other trackers to initialize HIT?}
However, for well-established tracking results, there is little room for further optimization,
resulting in limited improvements when directly applying HIT to them.
To tackle this issue, we propose to integrate HIT with other trackers in a simple \textit{recombination} manner.

Specifically, given raw tracking results $\mathcal{T}^0 = \{ \mathcal{T}^0_i \}_{i=1}^{N_0}$ from one tracker, 
we split each trajectory $\mathcal{T}^0_i$ at discontinuities to obtain multiple tracklets.
For example, if the frame indices are $[1,2,4,5,6]$, $\mathcal{T}^0_i$ will be split into two tracklets with indices $[1,2]$ and $[4,5,6]$ respectively.
This is based on the assumption that discontinuities in trajectories often indicate high unreliability.
Indeed, it is approximately equivalent to setting the maximum age of tracklets to 1 for the online tracker.
Afterwards, the new obtained tracklet set $\mathcal{T}^1 = \{ \mathcal{T}^1_i \}_{i=1}^{N_1} (N_0 \le N_1)$ is taken as the input of HIT.
In this context, HIT can be viewed as a post-processing method to refine the results from any tracker.
Fig.\ref{fig_integration} illustrates the overall integration pipeline.

\subsection{Discussions}

Currently, mainstream methods tend to rely on auxiliary cues (e.g., appearance, CMC), carefully tuned hyperparameters for each stage and sequence, 
and expensive training procedures to achieve outstanding performance.
Differently, HIT is designed as a heuristic method, which only uses IoU as association cues and sets unified hyperparameters for all hierarchies and sequences.
HIT can function both as an independent tracker and as a post-processing method for other trackers.
Introducing more information can certainly lead to better results.
Nevertheless, the main purpose of this paper is not to achieve the best performance, but rather to explore a concise and effective tracking framework.
We leave further optimizations for future work.

\begin{table*}[htbp]
    \begin{center}
        \resizebox{.95\textwidth}{!}{
            \begin{tabular}{lc|c|c|c|c|c|c|c|c|c}
                \toprule[1pt]
                & \textbf{Method} & \textbf{Param} & \textbf{Cues} & \textbf{Mode} 
                & \textbf{HOTA} & \textbf{DetA} & \textbf{AssA} & \textbf{MOTA} & \textbf{IDF1} & \textbf{IDSW} \\
                \\ [-2ex] \hline \\ [-2ex]
                & TransTrack \cite{sun2020transtrack}           & \checkmark &   L,O   & on  & 54.1 & 61.6 & 47.9 & 75.2 & 63.5 & 3,603 \\
                & TrackFormer \cite{meinhardt2022trackformer}   & \checkmark &    L    & on  & 57.3 & 60.9 & 54.1 & 74.1 & 68.0 & 2,829 \\   
                & MOTR \cite{zeng2022motr}                      & \checkmark &    L    & on  & 57.8 & 60.3 & 55.7 & 73.4 & 68.6 & 2,439 \\
                & MOTRv2 \cite{zhang2023motrv2}                 & \checkmark &   L,O   & on  & 62.0 & 63.8 & 60.6 & 78.6 & 75.0 & 2,619 \\
                & MOTRv3 \cite{yu2023motrv3}                    & \checkmark &    L    & on  & 60.2 & 62.1 & 58.7 & 75.9 & 72.4 & 2,403 \\
                & ByteTrack \cite{zhang2022bytetrack}           & \ding{55}  &   O,S   & on  & 62.8 & 63.8 & 62.2 & 78.9 & 77.2 & 2,310 \\
                &                                               & \ding{55}  &   O,S   & off & 63.2 & 64.4 & 62.3 & 79.7 & 77.4 & 2,253 \\
                & OC-SORT \cite{cao2023observation}             & \checkmark &   O,V   & on  & 61.7 & 61.6 & 62.0 & 76.0 & 76.2 & 2,199 \\
                &                                               & \checkmark &   O,V   & off & 63.2 & 63.2 & 63.4 & 78.0 & 77.5 & 1,950 \\ 
                & DeepSORT \cite{wojke2017simple}               & \checkmark &   O,A   & on  & 61.2 & 63.1 & 59.7 & 78.0 & 74.5 & 1,821 \\
                & StrongSORT \cite{du2023strongsort}            & \checkmark &  O,A,C  & on  & 63.5 & 63.6 & 63.7 & 78.3 & 78.5 & 1,446 \\
                &                                               & \checkmark &  O,A,C  & off & 64.4 & 64.6 & 64.4 & 79.6 & 79.5 & 1,194 \\
                & Hybrid-SORT \cite{yang2024hybrid}             & \ding{55}  &  O,V,S  & on  & 63.0 & 63.4 & 62.9 & 78.1 & 78.0 & 2,232 \\
                &                                               & \ding{55}  &  O,V,S  & off & 63.6 &  -   & 63.2 & 79.3 & 78.4 & 2,109 \\
                &                                               & \ding{55}  & O,V,S,A & on  & 63.2 & 63.5 & 63.1 & 78.4 & 78.2 & 1,296 \\
                &                                               & \ding{55}  & O,V,S,A & off & 64.0 &  -   & 63.5 & 79.9 & 78.7 & 1,191 \\
                & BoT-SORT \cite{aharon2022bot}                 & \ding{55}  &   O,C   & on  & 64.0 & 64.0 & 64.3 & 79.3 & 79.0 & 1,347 \\
                &                                               & \ding{55}  &   O,C   & off & 64.6 &  -   &  -   & \textbf{\textcolor{red}{80.6}} & 79.5 & 1,257 \\
                &                                               & \ding{55}  &  O,A,C  & on  & 64.3 & 63.9 & 64.9 & 79.4 & 79.4 & 1,353 \\
                &                                               & \ding{55}  &  O,A,C  & off & \textbf{\textcolor{red}{65.0}} & \textbf{\textcolor{red}{64.9}} 
                                                                                             & 65.5 & 80.5 & 80.2 & 1,212 \\
                & Deep OC-SORT \cite{maggiolino2023deep}        & \checkmark & O,V,A,C & on  & 63.3 & 62.1 & 64.9 & 76.6 & 79.1 & 1,146 \\
                &                                               & \checkmark & O,V,A,C & off & 64.9 & 64.1 & \textbf{\textcolor{red}{65.9}} 
                                                                                             & 79.4 & \textbf{\textcolor{red}{80.6}} & \textbf{\textcolor{red}{1,023}} \\
                & \textbf{HIT (ours)}                           & \textbf{\checkmark} &    \textbf{O}    & \textbf{off} 
                                                                & \textbf{63.5} & \textbf{64.2} & \textbf{63.2} & \textbf{79.3} & \textbf{77.4} & \textbf{1,461} \\
                \bottomrule[1pt]
            \end{tabular}
        }
        \caption{
            Performance comparison with state-of-the-art methods on MOT17 test set.
            ``Param'' indicates whether unified hyperparameters are used for all sequences.
            ``Cues'' represents the utilized information for association.
            The ``Mode'' column represents the tracking mode, where ``on'' stands for ``online'' and ``off'' stands for ``offline''.
            For hybrid methods that report results with offline post-processing tricks, we also reproduce the results of their online versions.
            Our HIT achieves higher HOTA / AssA and much lower IDSW than the offline version of ByteTrack, which tunes hyperparameters for all sequences.
        }
        \label{table_MOT17}
    \end{center}
\end{table*}

\section{Experiments}

\subsection{Experimental Setting}

\subsubsection{Datasets}

We conduct experiments on MOT17 \cite{milan2016mot16}, KITTI \cite{geiger2013vision}, DanceTrack \cite{sun2022dancetrack} and VisDrone \cite{zhu2018vision}. 
MOT17 is a widely used standard benchmark in MOT, which consists of 7 sequences, 5,316 frames for training and 7 sequences, 5,919 frames for testing.
For ablation, we split the training set into halves for training and validation as in previous works \cite{zhou2020tracking}.
KITTI is a popular dataset related to autonomous driving tasks, which consists of 21 training sequences and 29 test sequences with a relatively low frame rate of 10 FPS.
We use KITTI to validate the performance of HIT for tracking cars.
DanceTrack is a challenging dataset due to diverse non-linear motion patterns and severe occlusions.
It contains 40 sequences for training, 25 sequences for validation and 35 sequences for testing.
VisDrone is collected in UAV views, consisting of 56 training sequences, 7 validation sequences and 17 test-dev sequences.
Five object categories are considered for evaluation, i.e., car, bus, truck, pedestrian and van.

\subsubsection{Metrics}

We select HOTA \cite{luiten2021hota}, MOTA \cite{bernardin2008evaluating}, IDF1 \cite{ristani2016performance} and their related metrics for evaluation.
Specifically, MOTA focuses more on detection performance, IDF1 reflects the association capability, and HOTA balances these two aspects across various localization thresholds.

\subsubsection{Implementation Details}

For fair comparison, we directly use the detections from existing works.
For MOT17 and DanceTrack, we use the publicly available weights of YOLOX trained by ByteTrack \cite{zhang2022bytetrack}.
For KITTI, we borrow the results of PermaTrack \cite{tokmakov2021learning} following OC-SORT \cite{cao2023observation}.
For VisDrone, we use the trained YOLOX by U2MOT \cite{liu2023uncertainty}.
For association, a unified matching threshold $\Delta o = 0.2$ is utilized for all hierarchies, sequences and datasets.
The height modulated version of IoU \cite{yang2024hybrid} is utilized for person tracking on MOT17 and DanceTrack.
BYTE \cite{zhang2022bytetrack} is applied to include low-confidence detections.
The default width threshold $W$ in consistent-IoU is 64, and the consistent-camera threshold $\Delta O$ is set to 0.65.
For evaluation on test sets, extra interpolation and tracklets merging are performed similar with GIAOTracker \cite{du2021giaotracker}.
The hierarchical intervals are set to $\Delta t = [1, 5, 10, 15, 20, 30, \pm 5]$ as default, 
where ``±5'' means that a maximum overlap of 5 frames is allowed for association between tracklets in the last hierarchy.

\begin{table*}[htbp]
    \begin{center}
        \resizebox{.9\textwidth}{!}{
            \begin{tabular}{lc|c|c|c|c|c|c|c}
                \toprule[1pt]
                & \textbf{Method} & \textbf{HOTA} & \textbf{DetA} & \textbf{AssA} & \textbf{MOTA} & \textbf{FN} & \textbf{FP} & \textbf{IDSW} \\
                \\ [-2ex] \hline \\ [-2ex]
                & QDTrack \cite{fischer2023qdtrack} & 68.5 & - & 65.5 & 84.9 & - & - & 313 \\
                & IMMDP \cite{xiang2015learning} & 68.7 & 68.0 & 69.8 & 82.8 & 5,300 & 422 & 211 \\
                & AB3D \cite{weng20203d} & 70.0 & 71.1 & 69.3 & 83.6 & 11,836 & 2,305 & \textbf{\textcolor{red}{113}} \\
                & TuSimple \cite{choi2015near} & 71.6 & 72.6 & 71.1 & 86.3 & 3,656 & 759 & 292 \\
                & SMAT \cite{gonzalez2020smat} & 71.9 & 72.1 & 72.1 & 83.6 & 5,254 & \textbf{\textcolor{red}{175}} & 198 \\
                & TrackMPNN \cite{rangesh2021trackmpnn} & 72.3 & 74.7 & 70.6 & 87.3 & 2,577 & 1,298 & 481 \\
                & QD-3DT \cite{hu2022monocular} & 72.8 & 74.1 & 72.2 & 85.9 & 3,793 & 836 & 206 \\
                & CenterTrack \cite{zhou2020tracking} & 73.0 & 75.6 & 71.2 & 88.8 & 2,703 & 886 & 254 \\
                & LGMTracker \cite{wang2021track} & 73.1 & 74.6 & 72.3 & 87.6 & \textbf{\textcolor{red}{2,249}} & 1,568 & 448 \\
                & EagerMOT \cite{kim2021eagermot} & 74.4 & 75.3 & 74.2 & 87.8 & 3,497 & 454 & 239 \\
                & OC-SORT \cite{cao2023observation} & 74.6 & - & 74.5 & 87.8 & - & - & 257 \\
                & UCMCTrack \cite{yi2024ucmctrack} & 77.1 & - & 77.2 & 90.4 & - & - & - \\
                & PermaTrack \cite{tokmakov2021learning} & 77.4 & - & 77.7 & \textbf{\textcolor{red}{90.9}} & - & - & 275 \\
                & StrongSORT++ \cite{du2023strongsort} & \textbf{\textcolor{red}{77.7}} & \textbf{\textcolor{red}{77.9}} & 78.2 & 90.3 & 2,396 & 484 & 440 \\
                & \textbf{HIT (ours)} & \textbf{\textcolor{red}{77.7}} & 77.6 & \textbf{\textcolor{red}{78.3}} & \textbf{\textcolor{red}{90.9}} & \textbf{2,309} & \textbf{549} & \textbf{284} \\
                \bottomrule[1pt]
            \end{tabular}
        }
        \caption{
            Performance comparison on KITTI test set.
            Our HIT achieves similar tracking performance with previous state-of-the-art 2D tracking method StrongSORT++, especially with much lower IDSW.
        }
        \label{table_KITTI}
    \end{center}
\end{table*}

\begin{table*}[h]
    \begin{minipage}[htbp]{0.47\textwidth}
    \begin{center}
        \resizebox{1\textwidth}{!}{
            \begin{tabular}{lc|c|c|c|c|c}
                \toprule[1pt]
                & \textbf{Method} & \textbf{HOTA} & \textbf{DetA} & \textbf{AssA} & \textbf{MOTA} & \textbf{IDF1} \\
                \\ [-2ex] \hline \\ [-2ex]
                & FairMOT \cite{zhang2021fairmot} & 39.7 & 66.7 & 23.8 & 82.2 & 40.8 \\
                & TraDeS \cite{wu2021track} & 43.3 & 74.5 & 25.4 & 86.2 & 41.2 \\
                & TransTrack \cite{sun2020transtrack} & 45.5 & 75.9 & 27.5 & 88.4 & 45.2 \\
                & ByteTrack \cite{zhang2022bytetrack} & 47.7 & 71.0 & 32.1 & 89.6 & 53.9 \\
                & GTR \cite{zhou2022global} & 48.0 & 72.5 & 31.9 & 84.7 & 50.3 \\
                & MotionTrack \cite{qin2023motiontrack} & 52.9 & 80.9 & 34.7 & 91.3 & 53.8 \\
                & QDTrack \cite{fischer2023qdtrack} & 54.2 & 80.1 & 36.8 & 87.7 & 50.4 \\
                & MOTR \cite{zeng2022motr} & 54.2 & 73.5 & 40.2 & 79.7 & 51.5 \\
                & OC-SORT \cite{cao2023observation} & 55.1 & 80.3 & 38.3 & 92.0 & 54.6 \\
                & StrongSORT++ \cite{du2023strongsort} & 55.6 & 80.7 & 38.6 & 91.1 & 55.2 \\
                & PuTR \cite{liu2024putr} & 55.8 & - & - & 91.9 & 58.2 \\
                & MambaTrack$^+$ \cite{huang2024exploring} & 56.1 & 80.8 & 39.0 & 90.3 & 54.9 \\
                & GHOST \cite{seidenschwarz2023simple} & 56.7 & 81.1 & 39.8 & 91.3 & 57.7 \\
                & C-BIoU \cite{yang2023hard} & \textbf{\textcolor{red}{60.6}} & 81.3 & \textbf{\textcolor{red}{45.4}} & 91.6 & \textbf{\textcolor{red}{61.6}} \\
                & \textbf{HIT (ours)} & \textbf{56.6} & \textbf{\textcolor{red}{81.5}} & \textbf{39.5} & \textbf{\textcolor{red}{92.1}} & \textbf{55.4} \\
                \bottomrule[1pt]
            \end{tabular}
        }
        \caption{
            Performance comparison on DanceTrack test set.
            Our HIT achieves better results than offline method StrongSORT++ without heavy ReID and CMC components.
        }
        \label{table_DanceTrack}
    \end{center}
    \end{minipage}
    \hspace{9mm}
    \begin{minipage}[htbp]{0.47\textwidth}
    \begin{center}
        \resizebox{1\textwidth}{!}{
            \begin{tabular}{lc|c|c|c|c}
                \toprule[1pt]
                & \textbf{Method} & \textbf{HOTA} & \textbf{MOTA} & \textbf{IDF1} & \textbf{IDSW} \\
                \\ [-2ex] \hline \\ [-2ex]
                & MOTDT \cite{chen2018real} & - & -0.8 & 21.6 & 1,437 \\
                & SORT \cite{bewley2016simple} & - & 14.0 & 38.0 & 3,629 \\
                & MOTR \cite{zeng2022motr} & - & 22.8 & 41.4 & 959 \\
                & TrackFormer \cite{meinhardt2022trackformer} & - & 25.0 & 30.5 & 4,840 \\
                & GOG \cite{5995604} & - & 28.7 & 36.4 & 1,387 \\
                & UAVMOT \cite{liu2022multi} & - & 36.1 & 51.0 & 2,775 \\
                & DepthMOT \cite{wu2024depthmot} & 42.4 & 37.0 & 54.0 & 1,248 \\
                & AHOR-ReID \cite{jin2024ahor} & - & 42.5 & 56.4 & 810 \\
                & GIAOTracker \cite{du2021giaotracker} & 51.2 & 45.0 & 65.7 & \textbf{\textcolor{red}{616}} \\
                & STDFormer \cite{hu2023stdformer} & - & 45.9 & 57.1 & 1,440 \\
                & ByteTrack \cite{zhang2022bytetrack} & - & 52.3 & 68.3 & 2,230 \\
                & U2MOT \cite{liu2023uncertainty} & - & 52.3 & \textbf{\textcolor{red}{69.0}} & 1,052 \\
                & \textbf{HIT (ours)} & \textbf{\textcolor{red}{52.8}} & \textbf{\textcolor{red}{53.5}} & \textbf{65.6} & \textbf{909} \\
                \bottomrule[1pt]
            \end{tabular}
        }
        \caption{
            Performance comparison on VisDrone2019-MOT test-dev set.
            Our HIT shows comparable association performance (IDF1) with offline method GIAOTracker without utilizing appearance features.
        }
        \label{table_VisDrone}
    \end{center}
    \end{minipage}
\end{table*}

\subsection{Benchmark Results}

\subsubsection{MOT17}
We compare HIT with representative methods on MOT17 in Tab.\ref{table_MOT17}, and the best results are bolded in red.
For fair and clear comparison, except for commonly used metrics, we further add three columns, i.e., ``Param'', ``Cues'' and ``Mode''.
The ``\ding{55}'' in column ``Param'' means that hyperparameters are tuned for each sequence.
The ``Cues'' column indicates the information used for association.
In detail, ``O'' for overlap (e.g., IoU), ``S'' for score, ``V'' for velocity, ``A'' for appearance, ``C'' for CMC (i.e., camera movement compensation), and ``L'' for learning.
For the ``Mode'' column, ``on / off'' means online / offline tracking.
Specifically, for those online methods who report results with offline post-processing tricks,
we further reproduce and report their online tracking results.
It is shown that our HIT achieves better results than offline ByteTrack without tuning hyperparameters for each sequence or utilizing scores for association.

\subsubsection{KITTI}
As shown in Tab.\ref{table_KITTI}, HIT achieves the same HOTA and higer MOTA compared with the SOTA tracker StrongSORT++ on KITTI.
This is attributed to the hierarchical design of HIT, which enables reliable tracking in low frame-rate videos based on pure IoU.

\subsubsection{DanceTrack}
Tab.\ref{table_DanceTrack} presents the comparison between HIT and other trackers, and HIT surpasses many trackers.
There is still a gap compared to the SOTA method C-BIoU, because DanceTrack contains severe deformations and occlusions, reducing the reliability of IoU.

\subsubsection{VisDrone}
The comparison on VisDrone is shown in Tab.\ref{table_VisDrone}, and our HIT achieves the best MOTA and promising IDF1 metrics.
Specifically, HIT shows comparable association ability with GIAOTracker, which follows the hybrid pipeline and uses extra appearance features and CMC module.

\begin{table*}
    \begin{center}
        \resizebox{.8\textwidth}{!}{
            \begin{tabular}{lc|c|c|c|c|c|c|c|c|c|c}
                \toprule[1pt]
                & \multirow{2}*{\textbf{Method}}
                & \multicolumn{5}{c|}{\textbf{MOT17-val}} & \multicolumn{5}{c}{\textbf{KITTI-val}} \\
                \\ [-2ex] \cline{3-12} \\ [-2ex]
                & ~ & \textbf{HOTA} & \textbf{DetA} & \textbf{AssA} & \textbf{MOTA} & \textbf{IDF1} 
                    & \textbf{HOTA} & \textbf{DetA} & \textbf{AssA} & \textbf{MOTA} & \textbf{IDF1} \\
                \\ [-2ex] \hline \\ [-2ex]
                & Base (W) & 66.10 & 66.88 & 65.89 & 77.68 & 77.25 & 80.32 & 80.24 & 80.70 & 89.41 & 90.02 \\
                & Base (I) (ours) & \textbf{\textcolor{red}{67.22}} & \textbf{\textcolor{red}{67.10}} & \textbf{\textcolor{red}{67.88}} 
                                  & \textbf{\textcolor{red}{77.94}} & \textbf{\textcolor{red}{78.01}} 
                                  & \textbf{\textcolor{red}{81.03}} & \textbf{\textcolor{red}{80.24}} & \textbf{\textcolor{red}{82.11}} 
                                  & \textbf{\textcolor{red}{89.67}} & \textbf{\textcolor{red}{91.19}} \\            
                \bottomrule[1pt]
            \end{tabular}
        }
        \caption{
            Comparison between different hierarchical strategies, i.e., ``temporal window (W) ''and ``tracklet interval (I)''.
            Our method ``Base (I)'' obtains obvious better performance than previous hierarchical design ``Base (W)'', especially in association ability (AssA / IDF1).
        }
        \label{table_W_I}
    \end{center}
\end{table*}

\begin{table}[t]
    \begin{center}
        \resizebox{.45\textwidth}{!}{
            \begin{tabular}{lc|c|c|c|c|c|c|c}
                \toprule[1pt]
                & \textbf{Line} & \textbf{Method} & \textbf{CI} & \textbf{CC} & \textbf{CM} & \textbf{HOTA} & \textbf{MOTA} & \textbf{IDF1} \\
                \\ [-2ex] \hline \\ [-2ex]
                & 1 & Base &            &            &            & 67.22 & 77.94 & 78.01 \\
                & 2 &      & \checkmark &            &            & 67.38 & 77.82 & 78.32 \\
                & 3 &      &            & \checkmark &            & 67.64 & 78.58 & 78.57 \\
                & 4 &      &            &            & \checkmark & 67.37 & 78.25 & 78.29 \\
                & 5 &      & \checkmark & \checkmark &            & 67.88 & 78.41 & 79.07 \\
                & 6 &      & \checkmark &            & \checkmark & 67.71 & 77.74 & 78.95 \\
                & 7 &      &            & \checkmark & \checkmark & 67.70 & \textbf{\textcolor{red}{78.73}} & 78.72 \\
                & 8 &      & \checkmark & \checkmark & \checkmark & \textbf{\textcolor{red}{68.03}} & \textbf{78.12} & \textbf{\textcolor{red}{79.47}} \\
                \bottomrule[1pt]
            \end{tabular}
        }
        \caption{
            Ablation study of the three inconsistency solutions on MOT17 validation set.
            ``CI'', ``CC'' and ``CM'' are short for consistent-IoU, consistent-camera and consistent-motion.
            Our method improves baseline HOTA by 0.81, MOTA by 0.18 and IDF1 by 1.46.
        }
        \label{table_Ablation}
    \end{center}
\end{table}

\begin{table}[t]
    \begin{center}
        \resizebox{.37\textwidth}{!}{
            \begin{tabular}{lc|c|c|c|c}
                \toprule[1pt]
                & \textbf{Tracker} & \textbf{Post} & \textbf{HOTA} & \textbf{MOTA} & \textbf{IDF1} \\
                \\ [-2ex] \hline \\ [-2ex]
                & FairMOT &  -  & 57.32 & 69.14 & 72.66 \\
                &                                 & GSI & 58.83 & 71.08 & 73.74 \\
                &                                 & HIT & 57.34 & 69.19 & 72.54 \\
                &                                 & HIT$^\dagger$ & \textbf{\textcolor{red}{59.42}} & \textbf{\textcolor{red}{72.45}} & \textbf{\textcolor{red}{74.24}} \\
                \\ [-2ex] \hline \\ [-2ex]
                & TransTrack &  -  & 58.09 & 67.72 & 68.59 \\
                &                                     & GSI & 59.21 & 69.64 & 69.36 \\
                &                                     & HIT & 59.04 & 67.71 & 71.29 \\
                &                                     & HIT$^\dagger$ & \textbf{\textcolor{red}{61.12}} & \textbf{\textcolor{red}{71.66}} & \textbf{\textcolor{red}{73.15}} \\
                \\ [-2ex] \hline \\ [-2ex]
                & TrackFormer &  -  & 64.19 & 73.14 & 74.86 \\
                &                                             & GSI & 64.33 & 73.01 & 74.88 \\
                &                                             & HIT & 65.46 & 73.27 & 77.43 \\
                &                                             & HIT$^\dagger$ & \textbf{\textcolor{red}{66.25}} & \textbf{\textcolor{red}{74.40}} & \textbf{\textcolor{red}{78.01}} \\
                \\ [-2ex] \hline \\ [-2ex]
                & MOTDT &  -  & 65.29 & 75.48 & 76.34 \\
                &                           & GSI & 64.86 & 73.50 & 75.53 \\
                &                           & HIT & 67.28 & 77.02 & 79.46 \\
                &                           & HIT$^\dagger$ & \textbf{\textcolor{red}{67.78}} & \textbf{\textcolor{red}{77.87}} & \textbf{\textcolor{red}{79.80}} \\
                \\ [-2ex] \hline \\ [-2ex]
                & SORT &  -  & 66.32 & 74.73 & 77.62 \\
                &                              & GSI & 68.02 & 78.27 & 79.00 \\
                &                              & HIT & 66.22 & 74.77 & 77.80 \\
                &                              & HIT$^\dagger$ & \textbf{\textcolor{red}{68.33}} & \textbf{\textcolor{red}{79.05}} & \textbf{\textcolor{red}{79.51}} \\
                \\ [-2ex] \hline \\ [-2ex]
                & DeepSORT &  -  & 66.26 & 76.71 & 77.33 \\
                &                                 & GSI & 66.47 & 77.07 & 77.24 \\
                &                                 & HIT & 66.69 & 76.78 & 78.18 \\
                &                                 & HIT$^\dagger$ & \textbf{\textcolor{red}{67.69}} & \textbf{\textcolor{red}{78.90}} & \textbf{\textcolor{red}{78.98}} \\
                \\ [-2ex] \hline \\ [-2ex]
                & ByteTrack &  -  & 67.85 & 77.85 & 79.56 \\
                &                                     & GSI & 68.94 & 79.52 & 80.51 \\
                &                                     & HIT & 68.20 & 78.04 & 79.94 \\
                &                                     & HIT$^\dagger$ & \textbf{\textcolor{red}{69.44}} & \textbf{\textcolor{red}{80.27}} & \textbf{\textcolor{red}{80.95}} \\
                \bottomrule[1pt]
            \end{tabular}
        }
        \caption{
            Comparison experiments of post-processing methods on other trackers on the MOT17 validation set.
            ``GSI'' is Gaussian-smoothed interpolation,
            and ``HIT'' is our baseline method.
            ``$\dagger$'' represents applying interpolation and Gaussian smoothing. 
        }
        \label{table_integration}
    \end{center}
\end{table}

\subsection{Ablation Study}

\subsubsection{Hierarchical Strategy}

We explore the design of hierarchical strategy on MOT17 and KITTI validation sets in Tab.\ref{table_W_I}.
Previous hierarchical work \cite{Cetintas_2023_CVPR} utilizes sliding temporal windows (W) to partition different hierarchies.
Differently, our framework applies tracklet intervals (I) to determine the priority of associations.
Experimental results demonstrate that our method achieves consistent superiority over window-based method.
Particularly, the association metric AssA is improved by 1.99 and 1.41, respectively.

\subsubsection{Inconsistency Solutions}

The effects of three inconsistency solutions are investigated in Tab.\ref{table_Ablation}.
It is observed that:
\begin{itemize}
    \item Comparing line 1 and line 3, consistent-camera (CC) obviously improves all three metrics HOTA, MOTA and IDF1 without using any visual cues, 
    validating its effectiveness in countering camera movements.
    \item Comparing line 1 and line 4, consistent-motion (CM) improves MOTA and IDF1 by 0.31 and 0.28 respectively,
    which proves that it can enhance the association accuracy of the first hierarchy.
    \item Comparing line 7 and line 8, consistent-IoU (CI) can further improve IDF1 by 0.72.
    Please note that it harms MOTA because there also exists the ``inconsistent target size'' issue when computing IoU-based metrics, 
    and introducing CI results in an increase in FP (false positive) from 2,346 to 2,657.
    Even so, we retain this method because it can improve the overall tracking performance.
\end{itemize}

\subsubsection{Integration}

We integrate our baseline framework with seven other representative trackers,
including motion-based SORT and ByteTrack, appearance-based FairMOT, MOTDT and DeepSORT, and learning-based TransTrack and TrackFormer.
The results are show in Tab.\ref{table_integration}.
For comparison, interpolation method GSI \cite{du2023strongsort} is also included.
Obvious improvements over baseline trackers by introducing HIT can be observed, especially for IDF1.
Moreover, jointly applying HIT and interpolation (i.e., HIT$^\dagger$), 
the HOTA metrics exhibits an increase from 1.43 to 3.03.

\section{Conclusion}

In this paper, we present the hierarchical IoU tracking framework HIT, which performs hierarchical association based on tracklet intervals.
Experiments demonstrate its superiority over previous multi-stage or window-based methods. 
However, this pipeline faces three inconsistency issues, i.e., inconsistent target size, inconsistent camera movement, inconsistent hierarchical cues.
To solve these problems, we propose three corresponding solutions for more reliable associations.
Moreover, we prove that HIT can be integrated with any other trackers to refine the results, whether they are heuristic-based or learning-based.
Though it only relies on IoU for association, our HIT achieves promising performance on four datasets, i.e., MOT17, KITTI, DanceTrack and VisDrone,
proving its effectiveness and robustness.

However, there is still room for further performance improvements in challenging scenarios.
In future work, we will explore the integration of our hierarchical strategy and other optimizations.
For example, the simple IoU-based association can be replaced by elaborate learning-based modules.
We hope HIT can serve as a strong baseline for offline tracking and post-processing for future works. 

\newpage
\bibliography{aaai24}

\end{document}